\documentclass[conference]{IEEEtran}
\pdfoutput=1
\usepackage{eso-pic}
\usepackage{cite}
\usepackage{amsmath,amssymb,amsfonts}
\usepackage{algorithmic}
\usepackage{graphicx}
\usepackage{textcomp}
\def\BibTeX{{\rm B\kern-.05em{\sc i\kern-.025em b}\kern-.08em
    T\kern-.1667em\lower.7ex\hbox{E}\kern-.125emX}}
\begin{document}

\AddToShipoutPictureBG*{%
  \AtPageLowerLeft{%
\parbox[l]{\textwidth}{\small First published in the Proceedings of the 26th European Signal Processing Conference (EUSIPCO-2018) in 2018, published by EURASIP\\}
}}

\title{Reflection Analysis for Face Morphing Attack Detection}

\author{\IEEEauthorblockN{Clemens Seibold}
\IEEEauthorblockA{
\textit{Fraunhofer HHI}\\
Berlin, Germany \\
clemens.seibold@hhi.fraunhofer.de}
\and
\IEEEauthorblockN{Anna Hilsmann}
\IEEEauthorblockA{
\textit{Fraunhofer HHI}\\
Berlin, Germany \\
anna.hilsmann@hhi.fraunhofer.de}
\and
\IEEEauthorblockN{Peter Eisert}
\IEEEauthorblockA{
\textit{Fraunhofer HHI \& Humboldt University}\\
Berlin, Germany \\
eisert@informatik.hu-berlin.de}
}

\maketitle

\begin{abstract}
A facial morph is a synthetically created image of a face that looks similar to two different individuals and can even trick biometric facial recognition systems into recognizing both individuals. This attack is known as face morphing attack. The process of creating such a facial morph is well documented and a lot of tutorials and software to create them are freely available. Therefore, it is mandatory to be able to detect this kind of fraud to ensure the integrity of the face as reliable biometric feature. In this work, we study the effects of face morphing on the physically correctness of the illumination. We estimate the direction to the light sources based on specular highlights in the eyes and use them to generate a synthetic map for highlights on the skin. This map is compared with the highlights in the image that is suspected to be a fraud.
Morphing faces with different geometries, a bad alignment of the source images or using images with different illuminations, can lead to inconsistencies in reflections that indicate the existence of a morphing attack.

\end{abstract}

\begin{IEEEkeywords}
face morphing detection, reflection analysis, illumination estimation
\end{IEEEkeywords}

\section{Introduction}
\begin{figure}
[htbp]
\centerline{\includegraphics[width=0.5\textwidth]{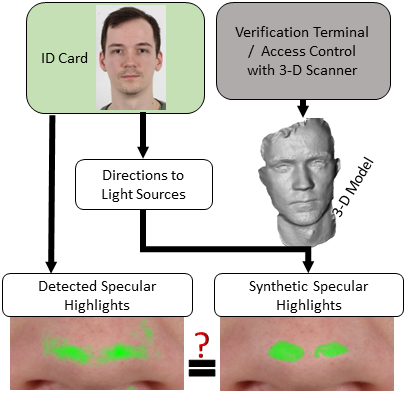}}
\caption{Overview of our morphing detection method}
\label{fig:Overview}
\end{figure}

Biometric verification systems are nowadays widely spread in commercial and sovereign applications like border control. Recently, the vulnerability of facial recognition systems against an specific attack was published by \cite{Ferrara}. They modified and fused face images of two different individuals such that the resulting face looks like booth individuals. They showed that even commercial facial recognition systems can be tricked by this attack and will match in an 1-to-1 matching mode both individuals with the forged face. 
Such an images is quite easy to create and tools for their generation are free available. The common practice is to take one image of each individual and align and blend them. The alignment is usually done by manually defining or automatically detecting, e.g. using dLib's \cite{dlib} implementation of the facial feature detector \cite{facialFeatDetector}, corresponding feature points in both faces. Both input images are warped such that the corresponding feature points are at the same position in both images. The aligned images are finally blended using an additive alpha blending. 
This kind of image forgery is known as morphing attack.\\
Several researchers become aware of the danger of morphing attacks and developed different forensic methods to detect this kind of fraud. In contrast to the already proposed methods, which are based on image degeneration \cite{Neubert17}, Binarized Statistical Image Features (BSIF) \cite{Raghavendra16}, neural networks \cite{Seibold17, Raghavendra17} or JPEG compression artifacts \cite{Makrushin17}, we propose a method that is based on a physical illumination model. 
Illumination estimation to detect frauds was already studied in detail by \cite{Kee10, Johnson07} to detect compositions of multiple photographs. They estimate the directions of the incident light individually for each object in an image and compared them to reveal forgeries. In case of morphing attacks, we have a different situation, since the whole image is usually manipulated and a good attacker would capture both input images under the same illumination. However, the alignment of faces with different geometry and slightly differing poses in the input images can lead to physically incorrect highlights on the skin that can be used to reveal frauds. To our knowledge, there is no morphing attack detection method based on physical models published yet.\\
Our method estimates the direction of the light sources in the suspicious image and synthesizes a reflection map using a skin reflection model, the 3-D geometry of the face and the estimated direction of the incident light. This reflection map is compared with the specular highlights in the suspicious image. In case of morphing attacks, the position and shape of the specular highlights may differ.
Figure \ref{fig:Overview} provides an overview of our fraud detection process for an application case. An individual wants to use an ID card for verification of his or her identity. The Verification Terminal, which needs to be equipped with a 3-D scanner, takes a 3-D scan of this individual, estimates the direction of the light sources in the image on the ID Card and detects the specular highlights in this image. Using the estimated light direction and the acquired 3-D model, the device synthesizes the specular highlights on the skin and compares them to the real highlights in the image on the passport.\\
We focus in this study on the specular reflections on the nose, since the nose is quite rigid compared to other regions in the face.
Our method can be divided into four different steps:
\begin{enumerate}
\item estimation of the direction of the incident light sources using specular highlights in the eyes
\item acquiring and fitting of the 3-D model to the suspicious image
\item synthesizes of highlights on the skin
\item comparison of synthetic highlights and highlights in the passport image
\end{enumerate}

In the next section, we describe all required estimations to be able to create a synthetic specular highlight map that is aligned with the suspicious image. In particular, we describe the estimation of the incident light direction, how we fit our 3-D model on the image to get surface normals for each pixel and the skin reflection model. Section 3 focuses on the setup of our experiments including the generation of morphing attacks and our capture setup. Finally, the synthetic and real highlights in genuine and forged images are presented and discussed.

\section{Estimation of Expected Highlights on the Skin}
The specular highlights in the eyes are a reliable source to get an estimate for the direction to the light source \cite{Johnson07}. The vitreous body has approximately the same shape and size for adults and since the view direction in passport images is specified (for ID cards usually even frontal), the highlights of two images are easy to align and no strong warping of the inner part of the eyes is necessary for the generation of a face morphing attack, if both input images are captured under the same condition. Thus, even the eyes of a morphed face image can be used to estimate the original illumination setup. Different setups would usually lead to different highlights in the eyes, which are difficult to align and often lead to visible artifacts. That is why a professional attacker would capture both images under the same conditions and keep the highlights in the eyes as they are.\\
In the following, we describe how we estimate the light source directions from the highlights in the eyes and how a map is synthesized that contains the highlights on the skin for a given 3-D model and illumination.
\subsection{Estimation of Light Directions}
\begin{figure}[htbp]
\centerline{
\includegraphics[width=0.45\textwidth]{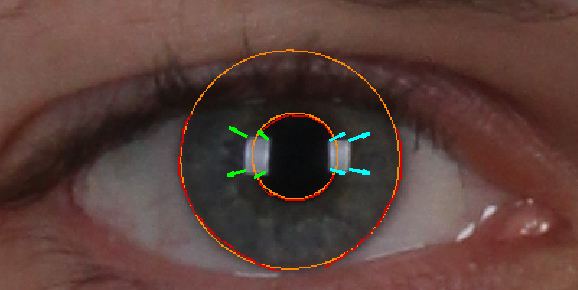}
}
\caption{Estimation of the light direction. The orange ellipses are the estimated border of the pupil and Limbus, the red points are the manually marked borders and the arrows illustrate the direction to the borders of the light sources.}
\label{fig:lightDirEst}
\end{figure}

The vitreous body of the human eye reflects light like a mirrored sphere. Its shape can be modeled by a sphere with a fixed radius of 7.8mm \cite{Lefohn03}. If we know the exact 3-D position of a vitreous body relative to the camera center, we also know its surface normals and can calculate for each visible highlight on the vitreous body the direction $\mathbf{L}$ to the light source that caused this highlight:
\begin{equation}
\mathbf{L} = 2(\mathbf{V}^\text{T}\mathbf{N})\mathbf{N}-\mathbf{V}
\end{equation}
with $\mathbf{N}$ being the surface normal of a point on the vitreous body and $\mathbf{V}$ being a normalized vector pointing from this point to the camera center. The vector $\mathbf{V}$ can be calculated using the focal length of the camera and the pixel position of the highlight in the image.
We estimate the direction of the incoming light for the light sources in each eye individually and average the position for each light source to get a more robust estimate. We neglect the difference in direction caused by the distance between the eyes, since it is only a few degree for a usual capture setup. \\ 
In order to get the 3-D position of the vitreous body, we employ the relation of the Limbus (the border between the cornea and the white of the eye) and the pupil relative to the vitreous body. The Limbus can be described by a circle with a radius of 5.8mm on the vitreous body \cite{Lefohn03}. The center of the pupil is located at the center of the Limbus with a small offset \cite{Wyatt95} and will be used to get a more robust estimate of the view direction. We use this eye model to estimate the position of the vitreous bodies in the image as follows.\\
First, we mark the Limbus and the border of the pupil manually in the image.
We initialize our model with an approximate position and orientation and afterwards refine it using a rigid body tracker \cite{eisert97}.
The initial distance from the camera to the center of the vitreous body can be calculated by
\begin{equation}
d = 5.8\text{mm} \cdot \frac{f}{r_l} - 5.25\text{mm}
\end{equation} 
with $f$ being the focal length and $r_l$ being the radius of the limbus in pixel. 5.25\text{mm} is the average distance between the center of the vitreous body and the limbus \cite{Lefohn03}. With given depth and the center of the limbus $[X,\ Y]$ in 2-D, we can approximate the 3-D position by calculating the corresponding 3-D point of $[X,\ Y]$: \begin{equation}
	[X,\ Y] \rightarrow \left[\frac{X}{f} \cdot d, \ \frac{Y}{f} \cdot d, \ d\right].
\end{equation}
The initial orientation of our model is set such that the eye looks towards the center of the camera. Our refinement consists of three steps that update the model parameters and create correspondences between points in the model and the marked pixels on the Limbus and border of the pupil.\\
First, we estimate the size of the pupil by calculating the average distance between the center of the Limbus and the pixels on the border pupil. Afterwards, we create correspondences between the 3-D points of the model and the 2-D points on the Limbus and pupil border. For each marked pixel, we look for the closest point in the model after projecting it into the image with the current parameters. Based on these correspondences, we update the orientation and position of the vitreous body using a rigid body tracker. These steps are repeated until the distance between the projected model and the marked pixels cannot be reduced anymore. Figure \ref{fig:lightDirEst} shows our eye model fitted into an image and the directions to the borders of the light sources. We used only one direction for each light source in our experiments, which was estimated for the center of the highlight, for simplicity and since the direction difference estimated for the same light source is negligible in our setup.

\subsection{Synthesis of Specular Highlights}
For the synthesis of the specular highlights, we need a 3-D model of the face, the light directions and a reflection model of the skin. In an application case, the 3-D model can be acquired by a 3-D scanner when applying for a passwort or when one claims the ownership of an ID card or passport at an electronic access control, for example, at an automated border control gate. The position of the light sources can be estimated as described above. To model the reflection function of the skin, we use the physically-based Torrance-Sparrow model \cite{TorranceSparrowModel}
\begin{eqnarray}
&f = \rho_s \frac{1}{\pi} \frac{D G}{(\mathbf{N}^\text{T} \mathbf{L})(\mathbf{N}^\text{T} \mathbf{V})}F(\mathbf{V}^\text{T} \mathbf{H})\\
\text{with }
&G = \min\left(1, \frac{2(\mathbf{N}^\text{T} \mathbf{H})(\mathbf{N}^\text{T} \mathbf{V})}{\mathbf{V}^\text{T} \mathbf{H}}, \frac{2(\mathbf{N}^\text{T} \mathbf{H})(\mathbf{N}^\text{T} \mathbf{L})}{\mathbf{V}^\text{T} \mathbf{H}}\right),\\
&D = \frac{1}{m^2(\mathbf{N}^\text{T}\mathbf{H})^4}\exp\left(-[(\tan (\mathbf{N}^\text{T}\mathbf{H}) / m)]^2\right),
\end{eqnarray} $\mathbf{H}$ being the half-way vector, $F$ the reflective Fresnel term and $\rho_s$ and $m$ parameters that depend on skin type, age and gender\cite{SkinRefModel}. In our experiments, we used $\rho_s = 0.497$, $m = 0.266$ and an relative refraction index of 1.38 for the Fresnel term as measured in \cite{SkinRefModel} for a male in his twenties with a white to olive skin type.\\
In order to synthesize a map of the highlights on the skin that is aligned to the passport image, we need the surface normals of the object at each pixel. For this purpose, we estimate the position and orientation of the 3-D model relative to the camera position of the passport image and project the normals into the image. The position and orientation is estimated in a analysis-by-synthesis approach. First, we render the 3-D model with the currently estimated parameters. Then, we calculate 2-D correspondences between points in our synthetic image and the image in the passport using dLib's facial feature detection \cite{dlib} and calculate an update for the position and orientation using a rigid body tracker \cite{eisert97}. These steps are repeated, until the average distance between the facial feature positions in the synthetic and passport image cannot be reduced anymore.

\section{Experiments}
\subsection{Morph Generation}
We generate our face morphs semiautomatically using the same approach as proposed in most current literature about morphing attacks. Actually, we estimate the position of facial features in each input image using dLib's facial feature detector, calculate the average position for each facial feature and apply a Delaunay triangulation on this set of averaged positions. The resulting mesh will serve as target mesh for our morphs. For a better quality of the morphs and to avoid visible artifacts due to estimation errors or complex geometry, we add and move some points manually.
The topology that we get from the Delaunay triangulation is applied on the point sets of facial features positions of the input images, so we have corresponding triangles between the input images and our target mesh. Now, we use an affine transformation on each region covered by a triangle in the input image to warp it to its corresponding region in the target mesh.\\
After this alignment procedure, we blend the input images using an additive alpha blending with an alpha of 0.5.
Since we are not interested in the whole face for our studies, but in the eyes and nose, we do not consider the background separately and blended the whole image, which will result in ghosting artifacts outside of the face. 
\begin{figure}[htbp]
\centerline{\includegraphics[width=0.16\textwidth]{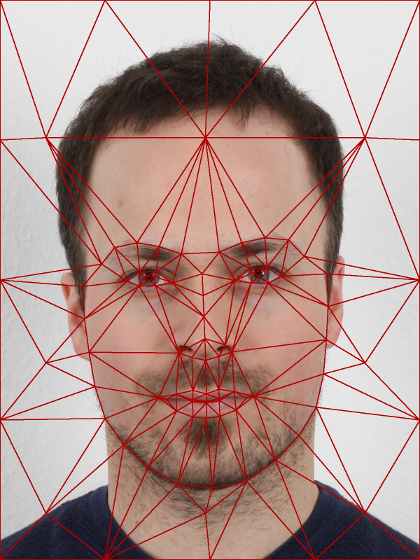}
\includegraphics[width=0.16\textwidth]{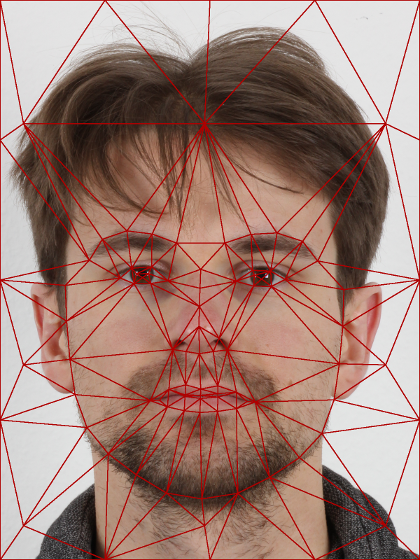}
\includegraphics[width=0.16\textwidth]{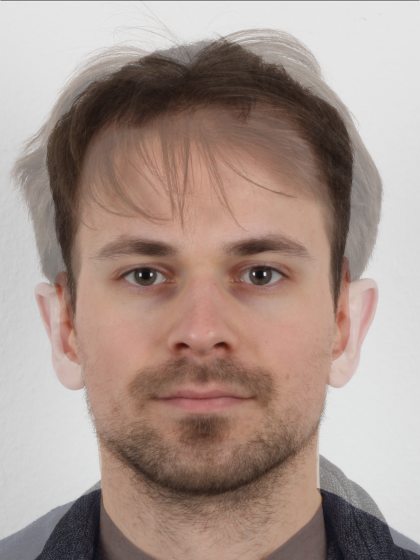}}
\caption{Example of a morphing attack.
The image on the left and in the center are the input images with the triangle mesh as overlay. The image on the right is the morphing attack based on these images with visible ghosting artifacts around the silhouette and hair.}
\label{fig}
\end{figure}

\subsection{Capture Setup}
We captured the 3-D models and the images used for testing in two different sessions with three months apart. We use two flashlights with soft boxes as diffusers, similar to the setup in a photographic studio. The cameras are mounted on a stand and the test person sits on a chair. That way, we get nearly the same head pose and thus position relative to the light sources and camera for each test person. The 3-D geometry is achieved using a passive optical flow based 3-D reconstruction approach \cite{Recon3DOF} that requires the use of a second camera. 

\section{Results}
\begin{figure}[htbp]
\center
\includegraphics[width=0.4\textwidth]{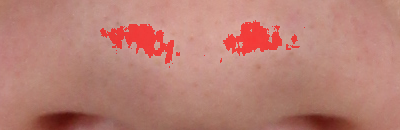}\\
\footnotesize{a) detected specular highlights}\\[1.5em]
\includegraphics[width=0.4\textwidth]{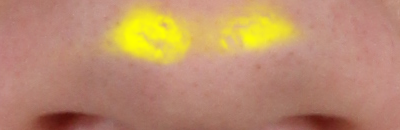}\\
\footnotesize{b) synthesized specular highlights}\\[1.5em]
\includegraphics[width=0.4\textwidth]{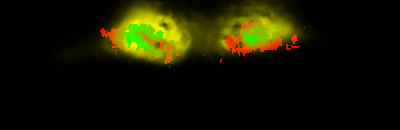}\\
\footnotesize{c) Detected (red) and synthesized (yellow) specular highlights, an overlap is visualized in green}
\caption{Detected and synthesized specular highlights in a genuine image}
\label{fig:Orig1}
\end{figure}

\begin{figure}[htbp]
\center
\includegraphics[width=0.4\textwidth]{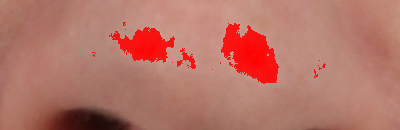}\\
\footnotesize{a) detected specular highlights}\\[1.5em]
\includegraphics[width=0.4\textwidth]{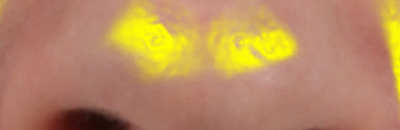}\\
\footnotesize{b) synthesized specular highlights}\\[1.5em]
\includegraphics[width=0.4\textwidth]{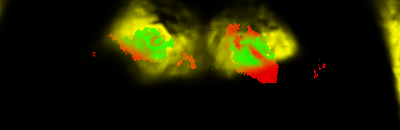}\\
\footnotesize{c) detected (red) and synthesized (yellow) specular highlights showing different shapes, an overlap is visualized in green}
\caption{Detected and synthesized specular highlights in a morphed face image}
\label{fig:Fake1}
\end{figure}

\begin{figure}[htb]
\vspace*{0.03in}
\center
\includegraphics[width=0.485\textwidth]{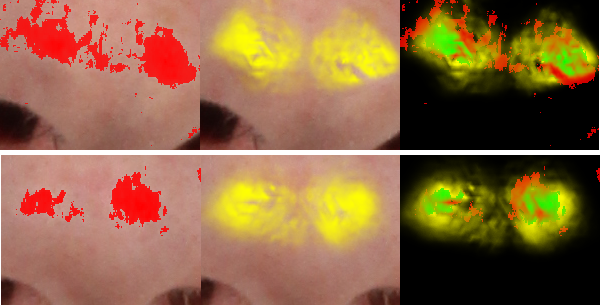}\\
\footnotesize{a) genuine images}\\[1.5em]
\includegraphics[width=0.485\textwidth]{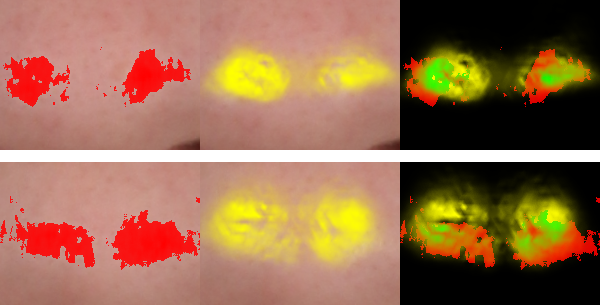}
\footnotesize{b) morphing attacks}
\caption{Examples of detected (left) and synthetic (center) specular highlights in genuine images and morphing attacks}
\label{fig:MoreExamples}
\end{figure}

We synthesize the specular highlights on the nose for several individuals as described above. Then, we compare them with the highlights in genuine images of the corresponding individuals and with morphing attacks that are created using an image of this individual. For the face morph generation and testing, we use only images that were taken three months apart from the images for the 3-D reconstruction.
The highlights in the images are detected using color saturation. We converted the image into the \textit{Hue Saturation Value color space} and define every pixel that has a saturation below a threshold as pixel in a specular highlight.\\
In the following, we show some representative examples of our highlight synthesis and detection for genuine images and morphing attacks. We selected them to illustrate the different kinds of existing highlights, whereby we skipped obvious frauds with a large difference of the highlight positions due to strong differences in the geometry of the faces.
Figure \ref{fig:Orig1} shows the synthetic and detected specular highlights of an genuine image. The shape of the synthetic and real highlight is similar and there is only a small shift in the location of the right highlight. The intensity reduction in the center of the left highlight is due to a rough surface in our 3-D model of this nose. This kind of abrupt and short reductions of specular highlights are also visible in nearly all of our synthetic highlights, see figure \ref{fig:Fake1} and \ref{fig:MoreExamples}, but affect only slightly their shape. The rough surface is caused by our passive 3-D capture method. We decided not to use smoothing methods since the depth-jump from the tip of the nose to the nostrils would lead to a strong distortive effect when employing smoothing operations.\\
The synthetic highlight in the morphing attack that is shown in figure \ref{fig:Fake1} has a different shape than the detected highlight. Whereas the synthetic highlight moves mostly horizontal, the detected highlight has a diagonal extent. Figure \ref{fig:MoreExamples} shows additional examples of synthetic and detected highlights for morphing attacks and genuine images. Again, the highlights in the genuine image are quite similar, whereas the highlights in the morphing attacks differ in shape and position. 

\section{Conclusion}
In this paper, we proposed a new morphing attack detection method based on a physical reflection model. We estimate the direction to light sources based on reflections in the eyes and synthesize the highlights on the nose using a 3-D model and a physically-based reflection model for the skin. The synthetic highlights that are created based on a physical reflection model, the 3-D geometry of a person and the orientation to the light sources differ in shape and location to the highlights that we get when using this person to create a morphing attack. In case of genuine images, the pattern of the synthesized and detected highlights are similar. As a result, we were able to distinguish between genuine images and morphing attacks.

\section*{Acknowledgment}
The work in this paper has been funded in part by the German Federal Ministry of Education and Research (BMBF) through the Research Program ANANAS under Contract No. FKZ: 16KIS0511.

\end{document}